\documentclass[preprint,12pt]{elsarticle}

\usepackage{amssymb}
\usepackage{amsmath,amssymb,amsfonts}
\usepackage{algorithm}
\usepackage{algpseudocode}
\usepackage{graphicx}
\usepackage{textcomp}
\usepackage{amssymb}
\usepackage{subfig}
\usepackage{graphicx}
\usepackage{url}
\usepackage{amsmath}
\usepackage{setspace}
\usepackage{multirow}
\usepackage{booktabs}

\begin{document}

\begin{frontmatter}



\title{Attention-based Fault-tolerant Approach for Multi-agent Reinforcement Learning Systems}

\author[1]{Mingyang Geng}
\author[1]{Kele Xu}
\author[1]{Yiying Li}
\author[2]{Shuqi Liu}
\author[1]{Bo Ding}
\author[1]{Huaimin Wang}

\address[1]{National Key Laboratory of Parallel and Distributed Processing, College of Computer, National University of Defense Technology, Changsha 410073, China; (e-mail: {gengmingyang13@nudt.edu.cn})}

\address[2]{National Key Laboratory of Big Data Management and Analysis, Northeastern University, Shenyang 110000, China. }

\begin{abstract}
The aim of multi-agent reinforcement learning systems is to provide interacting agents with the ability to collaboratively learn and adapt to the behavior of other agents. In many real-world applications, the agents can only acquire a partial view of the world. 
However, in realistic settings, one or more agents that show arbitrarily faulty or malicious behavior may suffice to let the current coordination mechanisms fail. In this paper, we study a practical scenario considering the security issues in the presence of agents with arbitrarily faulty or malicious behavior. Under these circumstances, learning an optimal policy becomes particularly challenging, even in the unrealistic case that an agent's policy can be made conditional upon all other agents' observations. To overcome these difficulties, we present an Attention-based Fault-Tolerant (FT-Attn) algorithm which selects correct and relevant information for each agent at every time-step. The multi-head attention mechanism enables the agents to learn effective communication policies through experience concurrently to the action policies. Empirical results have shown that FT-Attn beats previous state-of-the-art methods in some complex environments and can adapt to various kinds of noisy environments without tuning the complexity of the algorithm. Furthermore, FT-Attn can effectively deal with the complex situation where an agent needs to reach multiple agents' correct observation at the same time.

\end{abstract}

\begin{keyword}
Reinforcement learning \sep attention mechanism \sep fault tolerance


\end{keyword}

\end{frontmatter}


\section{Introduction}
\label{sec:introduction}

In this paper, we focus on the fault tolerance problem in multi-agent reinforcement learning systems. Consider the following robotic search and rescue scenario: a group of unmanned aerial vehicles (UAVs) is sent to find the survivors in a group of high-rise buildings after an earthquake \cite{geng2018learning}. The harsh environmental conditions might cause individual robots to fail, or hackers might take control of some robots and force them to behave in misleading ways \cite{Higgins2009survey}. In order to find the survivors as quickly as possible, these robots have to periodically exchange information with the neighbors and make decisions based both on the local view and the correct information from the neighbors. 

The above-mentioned multi-robot cooperation problem can be modeled as multi-agent reinforcement learning problem. Multi-agent reinforcement learning systems aim to provide interacting agents with the ability to collaboratively learn and adapt to other agents' behaviors. Plenty of real-world applications can be modeled as multi-agent systems, e.g. autonomous driving \cite{Dresner2008mutliagent}, smart grid control \cite{pipa2009multi} and multi-robot control \cite{geng2019learning}. Typically, an agent receives its private observations providing a partial view of the true state of the world. However, in realistic settings, one or more agents that show arbitrarily faulty or malicious behavior may suffice to let the current coordination mechanisms fail \cite{Millar2013towards}. Therefore, fault tolerance and credit assignment will become of paramount importance.

We conclude two challenges to enable the agents to collaboratively solve the underlying task with the fault-tolerance ability. First, a proper communication mechanism needs to be designed for the agents to extract correct and relevant information from others and model the environment. Then, the communication mechanism should maintain a stable complexity while keeping the ability to deal with different kinds of uncertainties in the environment i.e., accommodate a various number of agents with noisy observations without tuning the configuration of the algorithm. In concrete, the algorithm should maintain the ability to deal with the complex cases where an agent needs to reach multiple agents' correct observations at the same time.

To overcome the challenges, we present an Attention-based Fault-Tolerant (FT-Attn) algorithm which selects correct and relevant information for each agent at every time-step. The multi-head attention mechanism enables the agents to learn effective communication policies through experience concurrently to the main policies. Rather than simply sharing the correct observations, FT-Attn estimates the critic function for a single agent by selecting and utilizing the useful encoded information from others. We study the performance of FT-Attn in the modified Cooperative Navigation environments \cite{Kilinc18multi} and compare our results with the previous state-of-the-art method MADDPG-M (Multi-agent Deep Deterministic Policy Gradient-Medium) \cite{Kilinc18multi}. The results show a clear advantage of our method in some extremely complex environments. Furthermore, FT-Attn can easily adapt to various kinds of noisy environments without tuning the complexity of the algorithm. We also visualize the attention weights generated by FT-Attn to inspect how the fault-tolerance mechanism is working. We comment that FT-Attn is not designed for competing with other models without considering fault-tolerance, but a complementary one. We believe that adding our idea of fault-tolerance makes the existing algorithms much more valuable and practical.

The rest of this paper is organized as follows. Section \ref{sec:related_work} introduces the background and highly related work. 
Section \ref{sec:our_approach} describes the methodology of our work as well
as the architecture designed for training and prediction. The validation and evaluation of our work in the modified Cooperative Navigation environment are described in Section \ref{sec:experiments}. We conclude and provide
our future directions in Section \ref{sec:conclusion}. 




\section{Related Work}
\label{sec:related_work}

To the best of our knowledge, there is only one method MADDPG-M \cite{Kilinc18multi} studying the multi-agent reinforcement learning problem characterized by partial and extremely noisy observations, i.e., only one agent's observation is correct. To deal with the noisy observations which are weakly correlated to the true state of the environment, MADDPG-M forces the agents to learn whose private observation is sufficiently informative to be shared with others. However, the communication policy is task-specific relying on prior knowledge about the underlying environment requirement, which simplifies the uncertainties and the complexity compounded with a specific experimental evaluation. If the experimental setting changes, MADDPG-M must adjust the information filtering mechanism to adapt to the new environments. Besides, when there exist multiple correct observations (in not extremely noisy environments), MADDPG-M could not select the relevant information for each agent on the basis of correct observations and may lead to sub-optimal performance. Furthermore, the observation sharing mechanism may introduce redundant information (e.g. pixel data) because the raw observations may be high-dimensional.

\section{Our Approach}
\label{sec:our_approach}

In this section, we will first introduce the problem formulation of our multi-agent reinforcement learning fault-tolerant setting. Then, we will introduce the framework of our proposed method FT-Attn. Finally, we will introduce the training details.

\subsection{Problem Formulation}

We consider partially observable Markov Games, and make the assumption that the observations received by some of the agents are noisy and weakly correlated to the true state, which makes learning optimal policies unfeasible. Denote the policy for agent $i$ on all $N$ private observations as $a_{i}=\mu_{i}(o_{1},\cdots,o_{n})$. The learning process of the individual policy $a_{i}$ is hard to complete because a large number of $o_{i}$ are uncorrelated to the corresponding true state $s$, i.e., the background information provides a poor representation of the current true state for the $i_{th}$ agent. In order to solve this challenge, each agent has to explicitly and selectively exploit the correct and useful observations shared by other agents. In other words, the agents have to form a common cognition internally before they master the ability to cooperate. Due to the reason that the agents cannot discriminate between relevant and noisy information on their own, the ability to decide whether to share their own observations with others must also be acquired through experience. 

\subsection{Framework of FT-Attn}

More formally, we introduce multi-head attention mechanism to learn the critic for each agent by selectively paying attention to other agents' observations. Figure 1 illustrates the main components of our approach.

\begin{figure*}
	\centering
	\includegraphics[width=0.8\textwidth]{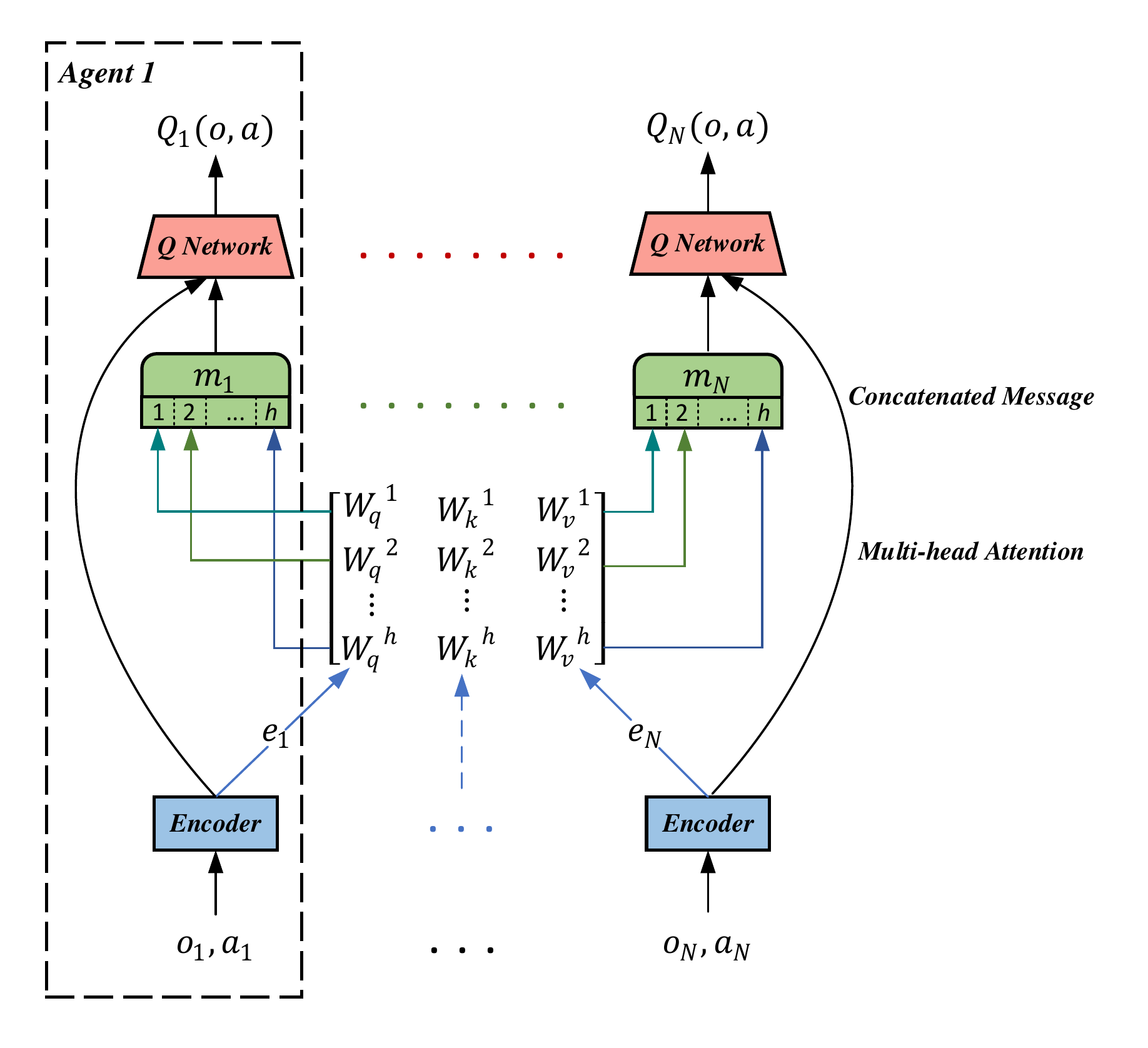}
	\caption{FT-Attn is composed of three modules: encoder, multi-head attention part, and Q network.}
	\label{fig1}
\end{figure*}

We use multi-head dot-product attention to select the correct observations and compute interactions between agents. Intuitively, each agent inquires about other agents for information about their observations as well as the actions and then takes the relevant information into account for estimating its value function.  Denote $Q_{i}^{\psi}(o,a)$ as the function of agent $i$'s observation and action, as well as other agents' contributions, the value is estimated as follows:

\begin{equation}
Q_{i}^{\psi}(o,a)=f_{i}(g_{i}(o_{i},a_{i}),m_{i}).
\end{equation}

Here, $f_{i}$ represents the Q-Network and $g_{i}$ represents the encoder function. The contribution from other agents $m_{i}$ is a weighted sum of each agent's value:

\begin{equation}
m_{i}=\sigma\left(\text { Concat }\left[\sum_{j \in \backslash i} \alpha_{i j}^{h} \mathbf{W}_{v}^{h} e_{j}, \forall h \in \mathbb{H}\right]\right),
\end{equation}
where $h$ is an attention head and $e_{j}$ is the embedding encoded by $g_{j}$ function. $W_{v}^{h}$ transforms $e_{j}$ into a ``value''. The set of all agents except $i$ is represented as $\backslash i$ and indexed with $j$. To calculate the weight $a_{ij}^{h}$, the input feature of each agent is projected to query, key and value representation by each independent attention head. For attention head $h$, the relation between $i$ and $j$ is computed as:

\begin{equation}
\alpha_{i j}^{h}=\frac{\exp \left(\tau \cdot \mathbf{W}_{q}^{h} e_{i} \cdot\left(\mathbf{W}_{k}^{h} e_{j}\right)^{\top}\right)}{\sum_{r \in \backslash i} \exp \left(\tau \cdot \mathbf{W}_{q}^{h} e_{i} \cdot\left(\mathbf{W}_{k}^{h} e_{r}\right)^{\top}\right)},
\end{equation}
where $\tau$ is a scaling factor, $W_{q}^{h}$ transforms $e_{i}$ into a ``query'' and $W_{k}^{h}$ transforms $e_{j}$ into a ``key''. 

\subsection{Training Details of FT-Attn}

All critics are updated together to minimize a joint regression loss function because of the parameter sharing:

\begin{equation}
L_{Q}(\psi)=\sum_{i=1}^{n}E_{(o,a,r,o^{'})\sim D}[(Q_{i}^{\psi}(o,a)-y_{i})^{2}], 
\end{equation}  
where
\begin{equation}
y_i = r_i+\gamma E_{\alpha^{'}\sim{\pi\bar{\theta}}(o^{'})}[Q_{i}^{\bar{\psi}}(o^{'},a^{'})-\alpha log(\pi_{\bar{\theta_i}}(a_i^{'}|o_i^{'}))],
\end{equation}  
where $\bar{\psi}$ and $\bar{\theta}$ are the parameters of the target critics and target policies respectively. $\alpha$ represents the temperature parameter determining the balance between maximizing entropy and rewards \cite{li18attention}. The individual policies are updated with the following gradient:

\begin{equation}
\bigtriangledown_{\theta_i}J(\pi_\theta)=E_{\alpha\sim\pi_\theta}[\bigtriangledown_{\theta_i}log(\pi_{\theta_{i}}(a_i|o_i))(\alpha log(\pi_{\theta_{i}}(a_i|o_i))-Q_i^{\psi}(o,a)+b(o,a_{\backslash i})],
\end{equation}
where $b(o,a_{\backslash i})$ represents the multi-agent baseline used to calculate the advantage function:

\begin{equation}
b(o,a_{\backslash i})=E_{a_{i}\sim{\pi_{i}(o_{i})}}[Q_{i}^{\psi}(o, (a_{i}, a_{\backslash i}))]=\sum_{a_{i}^{'}\in{A_{i}}}\pi(a_{i}^{'}|o_{i})Q_{i}(o,(a_{i}^{'},a_{\backslash i})),
\end{equation}
the advantage function here can help solve the multi-agent credit assignment problem \cite{Foerster2018counter}. In concrete, by comparing the value of specific action to the value of the average action for the agent, with all other agents fixed, we can know whether any increase in reward is attributed to other agents' actions.

For the training procedure, we use Soft Actor-Critic \cite{Haarnoja2018Soft} method for maximum entropy reinforcement learning. The model is trained for $10^{5}$ episodes with 25 steps each episode. We add a tuple of $(o_{t},a_{t},r_{t},o_{t+1})_{1\cdots N}$ to the replay buffer with the size of $10^{6}$ at each time-step. We update the network parameters after every 1024 tuples added to the replay buffer and perform gradient descent on the loss function. We use Adam \cite{Kingma2014adam} optimizer with a learning rate of 0.001. For other hyper-parameters, the discount factor $\gamma$ is set to 0.99; the dimension of the hidden state is set to 128 and the number of attention heads is set to 4. For the exploration noise, following \cite{Li15conti}, we use an Ornstein-Uhlenbeck process \cite{Uhlenbeck1930on} with $\theta=0.15$ and $\sigma=0.2$.

\section{Experiments}
\label{sec:experiments}

In this section, we will first introduce the experimental setting and baseline methods. Then, we will show the experimental results compared with the baseline methods. Finally, we will give the attention visualization and the corresponding analyze.

\subsection{Experimental Setting and Baseline Methods}

To evaluate our proposed approach, we follow the experimental settings: three varied versions of Cooperative Navigation problem ($N$ agents and $N$ landmarks) in MADDPG-M \cite{Kilinc18multi}. In concrete, only one gifted agent of $N$ agents can observe the true position of the landmarks and all other agents receive inaccurate information about the landmarks' positions. The task includes three different variants of increasing complexity depending on how the gifted agent is defined: in the Fixed case, the gifted agent stays the same throughout the training phase; in the Alternating case, the gifted agent may change at the beginning of each episode; in the Dynamic case, the agent closest to the center of the map becomes the gifted one within each episode. 

We set $N=3$ and evaluate FT-Attn against five actor-critic based baselines, DDPG (Deep Deterministic Policy Gradient) \cite{Li15conti}, MADDPG (Multi-agent Deep Deterministic Policy Gradient) \cite{Lowe17multi}, Meta-agent \cite{Kilinc18multi}, DDPG-OC (DDPG with Optimal Communication) \cite{Kilinc18multi} and MADDPG-M \cite{Kilinc18multi}. DDPG and MADDPG are chosen to provide a lower bound of the performance since each agent's policy is conditioned only on its own observations. The Meta-agent method is included to test the performance where all the observations are shared. The DDPG-OC baseline is chosen to study what level of performance is achievable when communicating optimally. MADDPG-M enables concurrent learning of optimal communication policy and the underlying task.

\subsection{Performance Comparison with the Baseline Methods}

We run an additional 1000 episodes after training to collect the performance metrics and report the averages on the three scenarios. Table 1 shows the mean episode rewards for FT-Attn and all baselines on the three scenarios. In these cases, both DDPG and MADDPG fail to learn the correct behavior because the observations are not allowed to be shared in the execution process. DDPG-OC provides an upper bound of the performance since the shared message is correctly controlled. The poor performance provided by Meta-agent method in the complex cases (Alternating and Dynamic) demonstrates that selecting correct and relevant information is essential in noisy environments. FT-Attn beats MADDPG-M in the complex cases and performs quite similarly to the upper bound DDPG-OC. The superiority demonstrates that multi-head attention mechanism maintains a better ability to deal with the noisy observations.

\begin{table*}[htbp] 
	\centering
	\caption{\label{tab:test}Mean (standard deviations) episode rewards for all baselines in all 3 scenarios.} 
	\begin{tabular}{cccc} 
		\toprule 
		Approach & Fixed & Alternating & Dynamic \\ 
		\midrule
		Meta-agent & -39.95 $\pm$ 4.50 & -51.42$\pm$ 7.70 & -60.98 $\pm$ 8.82 \\ 
		DDPG-OC & -39.26 $\pm$ 4.45 & -43.44$\pm$ 5.92 & -41.25 $\pm$ 5.24 \\
		MADDPG & -54.00 $\pm$ 7.43 & -58.67$\pm$ 8.90 & -63.44 $\pm$ 9.88 \\ 
		DDPG & -56.00 $\pm$ 8.96 & -56.50$\pm$ 8.51 & -60.66 $\pm$ 8.68 \\
		MADDPG-M & -39.73 $\pm$ 5.09 & -43.34 $\pm$ 7.29 & -43.91 $\pm$ 7.75 \\ 
		\textbf{FT-Attn} & \textbf{-40.89 $\pm$ 5.42} & \textbf{-42.75 $\pm$ 7.83} & \textbf{-42.48 $\pm$ 7.36} \\
		\bottomrule 
	\end{tabular} 
\end{table*} 

\subsection{Attention Visualization}

To understand how the use of attention contributes to the fault-tolerance ability, we examine the ``entropy'' of the four attention weights from the attention heads in Figure 2, Figure 3 and Figure 4. The initial value $0.69$ represents the maximum possible entropy (i.e. uniform attention across all agents). Lower entropy indicates that the head is focusing on specific agents. We find that the agents are more willing to utilize head1 and head2 and each agent appears to use a different combination of the four heads.

\begin{figure*}
	\centering
	\includegraphics[width=0.8\textwidth]{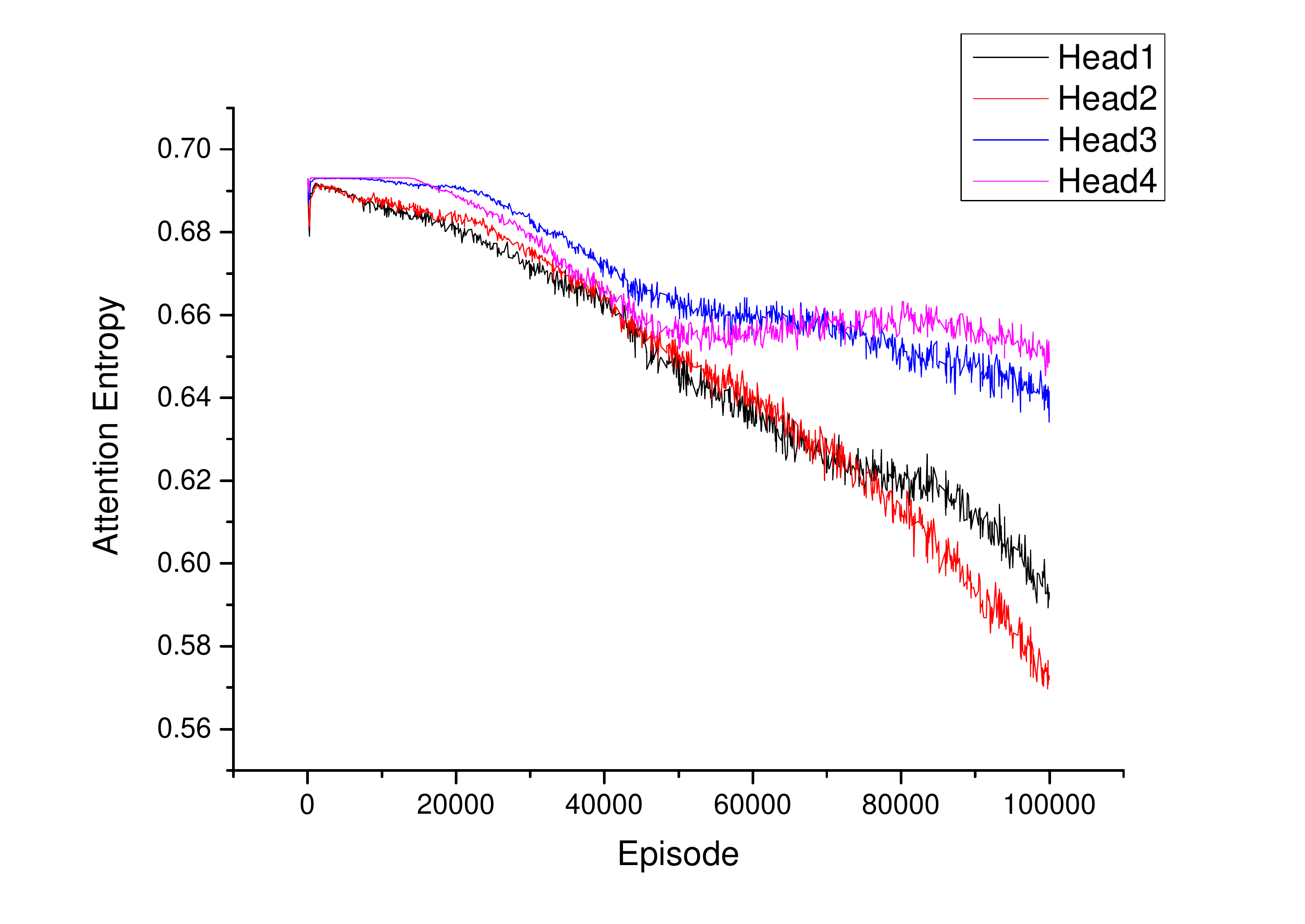}
	\caption{Attention entropy for each head over the course of training for agent 1 in the ``Dynamic'' situation.}
	\label{fig1}
\end{figure*}

\begin{figure*}
	\centering
	\includegraphics[width=0.8\textwidth]{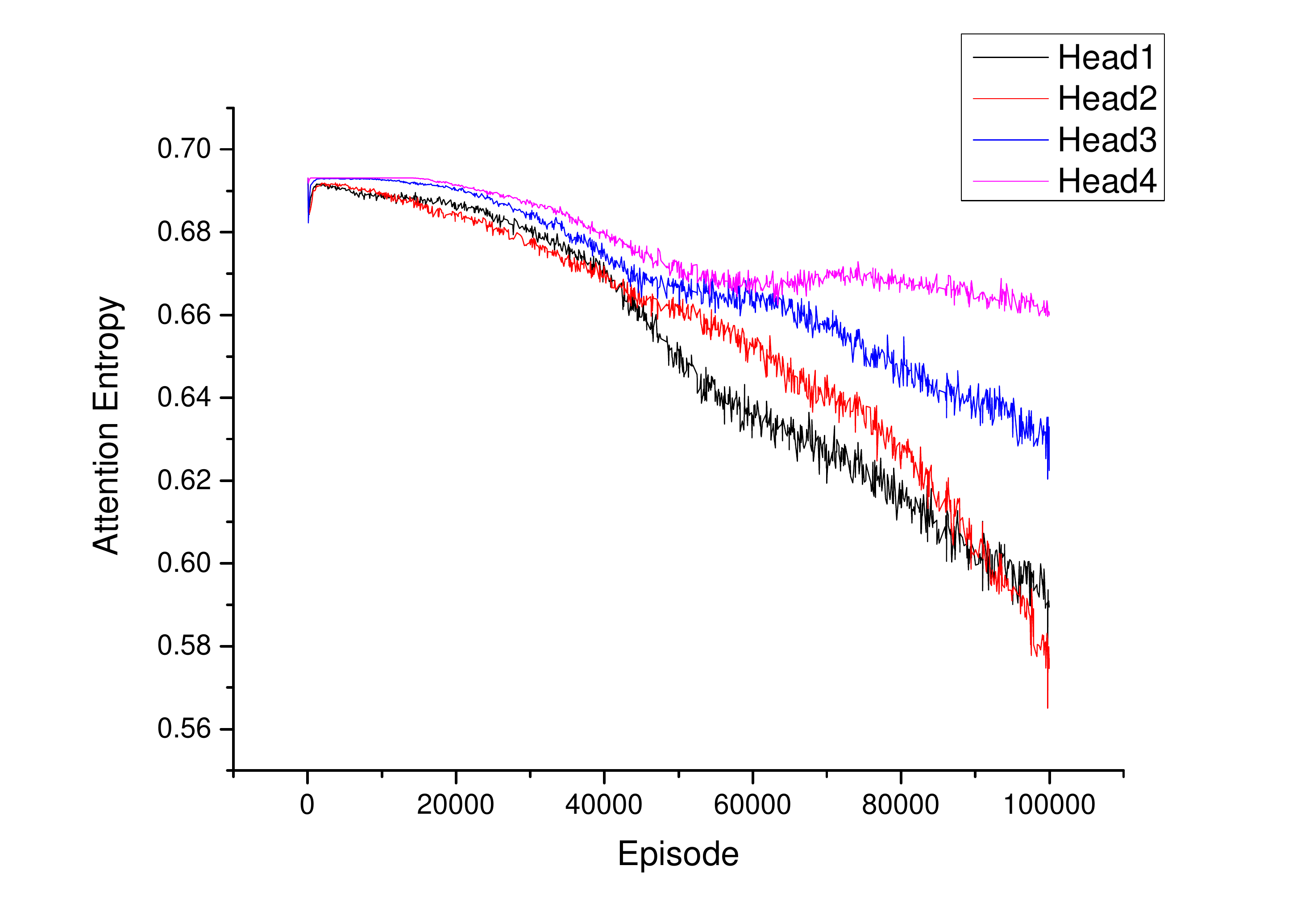}
	\caption{Attention entropy for each head over the course of training for agent 2 in the ``Dynamic'' situation.}
	\label{fig1}
\end{figure*}

\begin{figure*}
	\centering
	\includegraphics[width=0.8\textwidth]{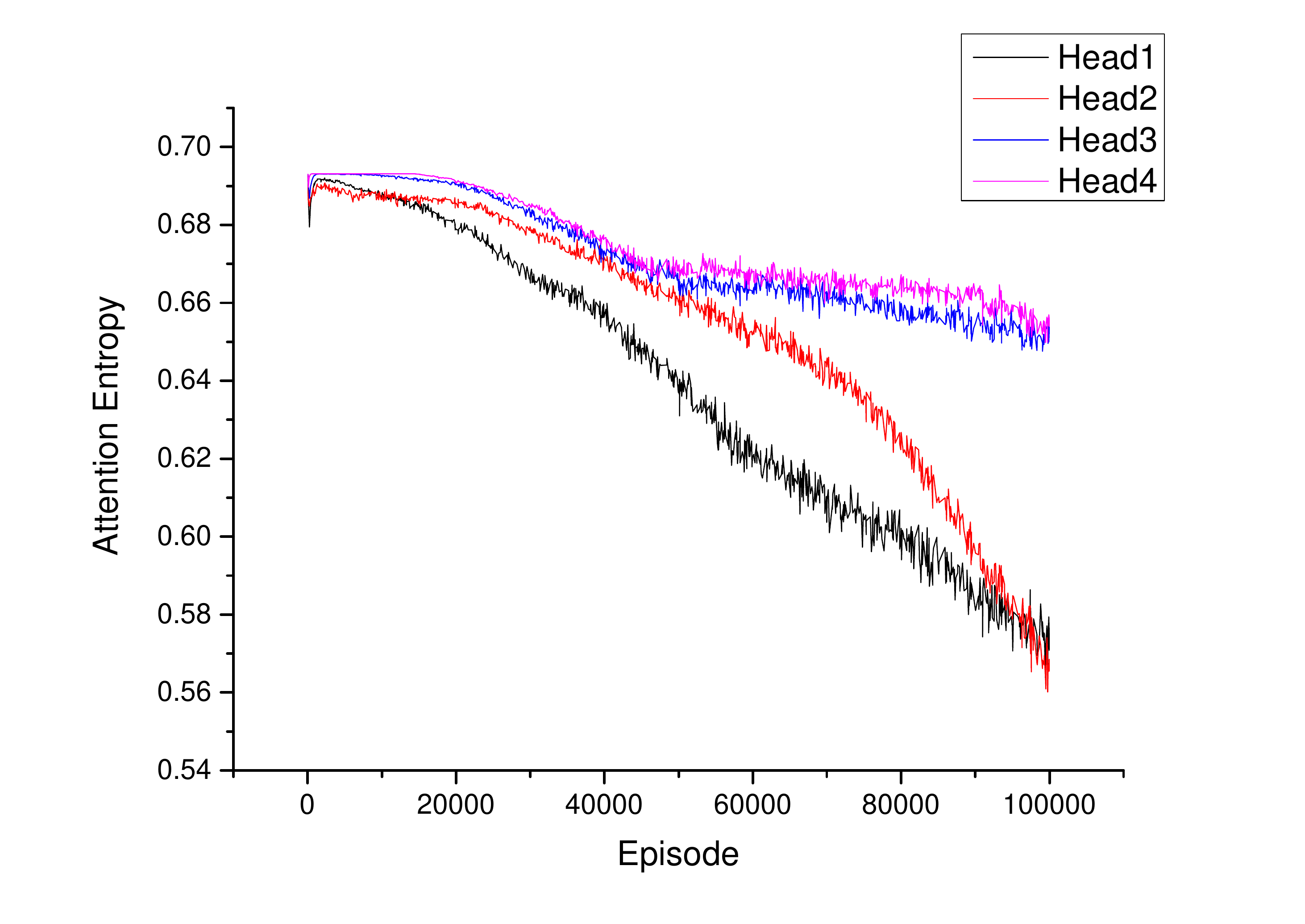}
	\caption{Attention entropy for each head over the course of training for agent 3 in the ``Dynamic'' situation.}
	\label{fig1}
\end{figure*}

We further visualize the attention weights generated by FT-Attn to understand the interactions in $N=5$ scenarios containing a different number of gifted agents. Each agent's attention weight is calculated from the heads that the agent appears to use the most. We pick the scenarios containing four different combinations of the gifted agents (with the number of gifted agents set to 1, 2, 3 and 4 respectively) and show the related heat-maps of the interaction matrix generated by FT-Attn in Figure 5 and Figure 6. We can see that the agents have acquired the ability to select the correct observations (self-attention is avoided in FT-Attn). Furthermore, the agents can also select the relevant and useful information among the correct observations (not paying uniform attention), which demonstrates the superiority compared with the simple information sharing mechanism in MADDPG-M.

\begin{figure}[htbp]
	
	\centering
	\subfloat[]{
		\begin{minipage}{5cm}
			\centering
			\includegraphics[width=5.5cm]{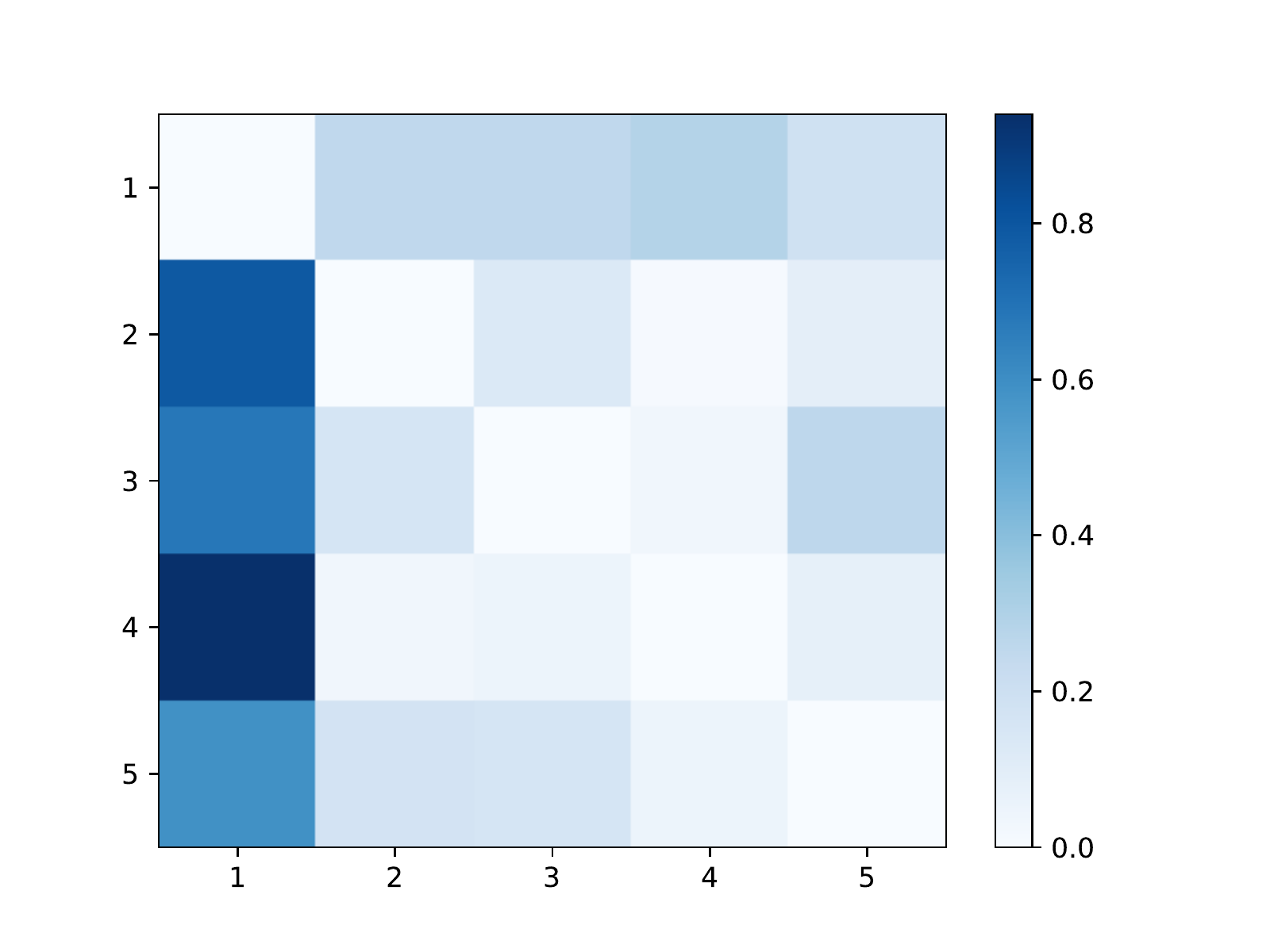}
		\end{minipage}%
	}%
	\subfloat[]{
		\begin{minipage}{5cm}
			\centering
			\includegraphics[width=5.5cm]{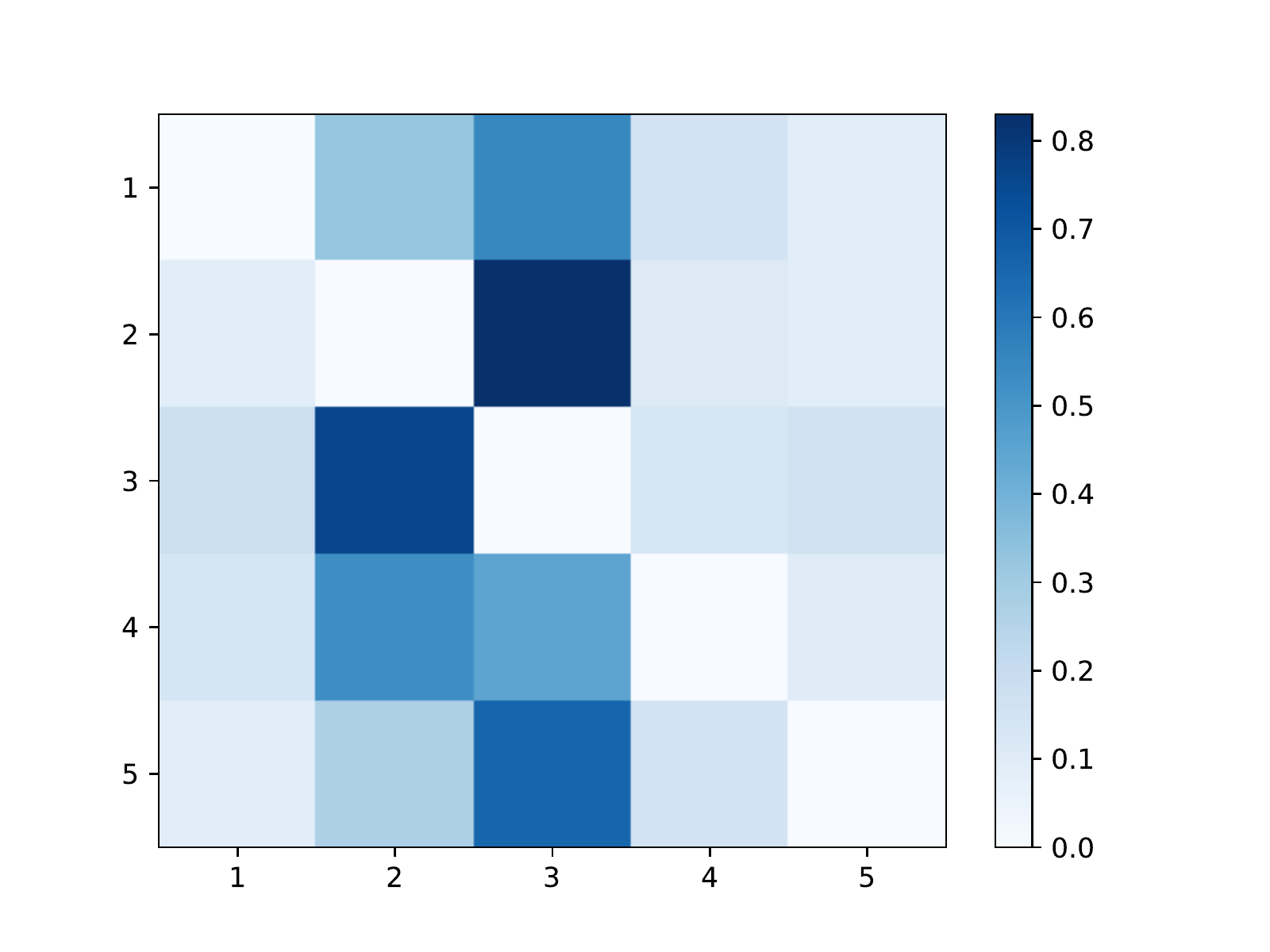}
		\end{minipage}%
	}%
	
	\caption{Attention weights generated by FT-Attn in the Fixed case of $N=5$ version. (a) Scenario 1: only the observation of agent 1 is correct; (b) Scenario 2: the observations of agent 2 and agent 3 are correct.}
	
\end{figure}

\begin{figure}[htbp]
	
	\centering
	\subfloat[]{
		\begin{minipage}{5cm}
			\centering
			\includegraphics[width=5.5cm]{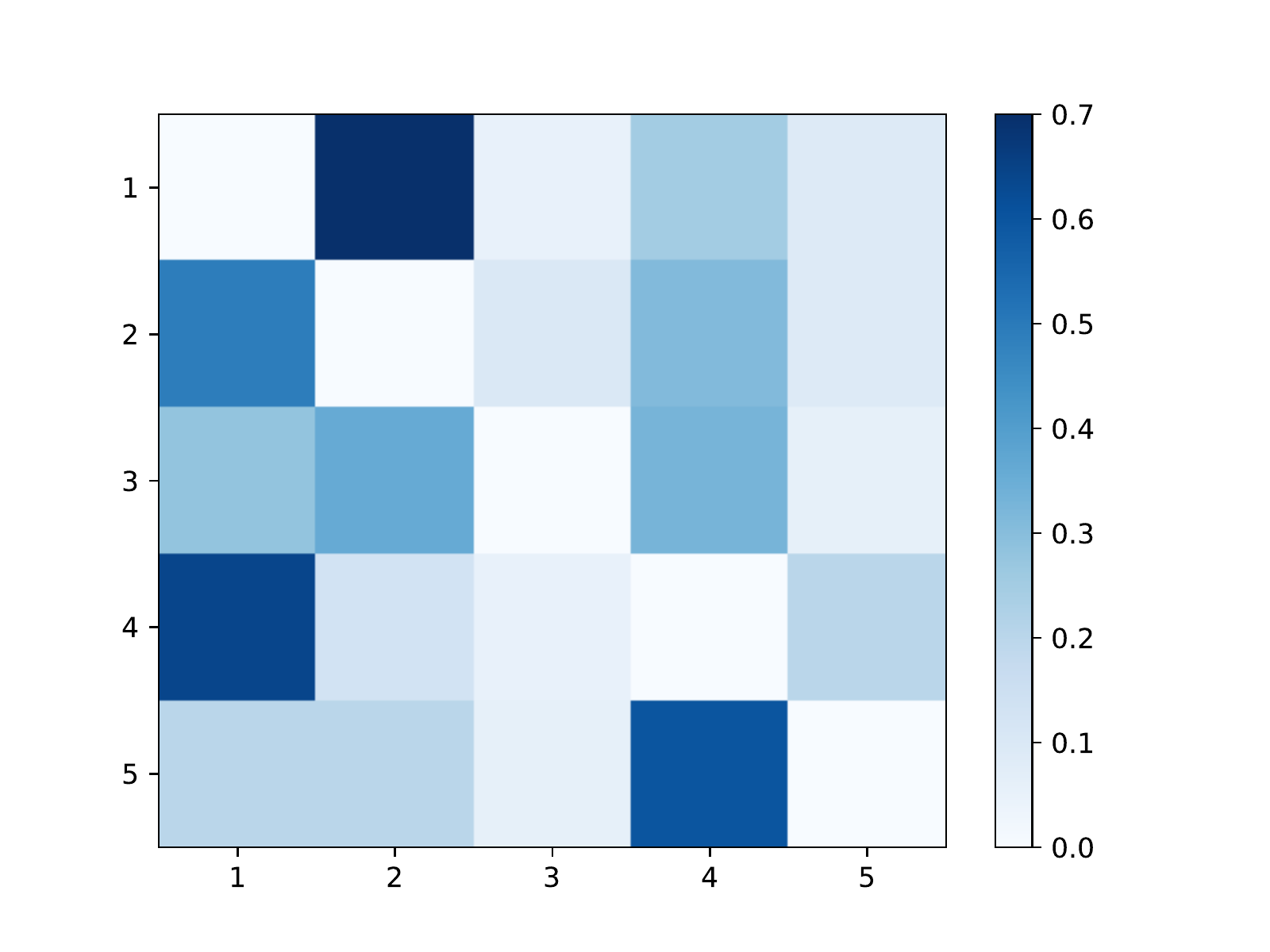}
		\end{minipage}%
	}%
	\subfloat[]{
		\begin{minipage}{5cm}
			\centering
			\includegraphics[width=5.5cm]{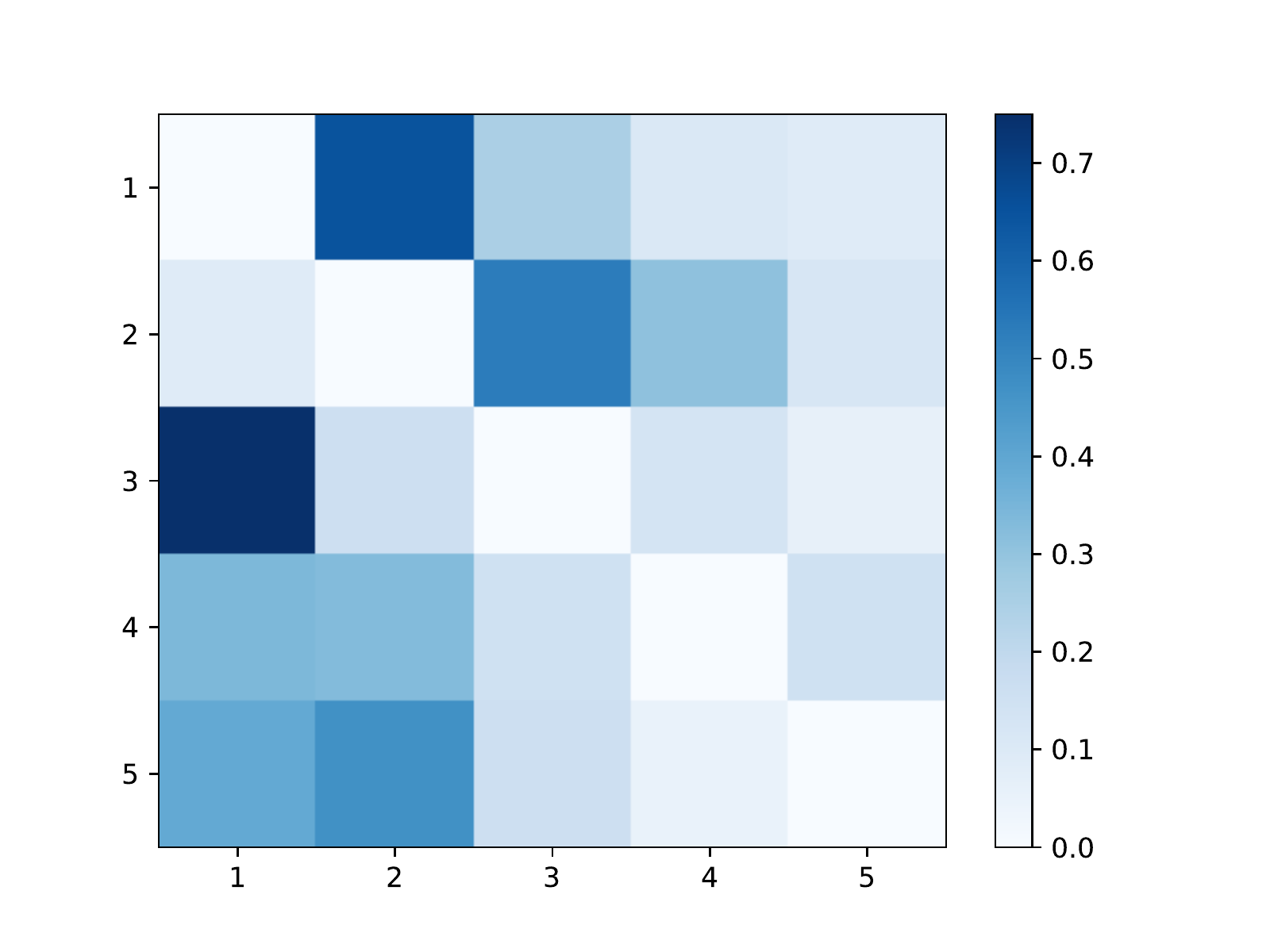}
		\end{minipage}%
	}%
	
	\caption{Attention weights generated by FT-Attn in the Fixed case of $N=5$ version. (a) Scenario 3: the observations of agent 1, agent 2 and agent 4 are correct; (b) Scenario 4: the observations of agent 1, agent 2, agent 3 and agent 4 are correct.}
	
\end{figure}

\section{Conclusion}
\label{sec:conclusion}

We propose an algorithm FT-Attn for coping with the fault-tolerance problem in the multi-agent reinforcement learning systems. The key idea is to utilize the multi-head attention mechanism to select the correct and useful information for estimating critics. We evaluate the performance of FT-Attn in the modified Cooperative Navigation environments compared with MADDPG-M, a method which is designed for dealing with extremely noisy environments. Empirical results have shown that FT-Attn beats MADDPG-M in some complex environments and can adapt to various kinds of noisy environments without tuning the complexity of the algorithm. Furthermore, FT-Attn can effectively deal with the complex situation where an agent needs to reach multiple agents' correct observation at the same time. 

In our future work, we will compare the performance of FT-Attn with other baseline methods in Predator and Prey scenario. Besides, we will increase the number of agents and further highlight the advantage of fault-tolerance ability in multi-agent reinforcement learning systems. 

\section*{Acknowledgment}

This work was supported by the National Natural Science Foundation of China (Grant Numbers 61751208, 61502510, and 61773390), the Outstanding Natural Science Foundation of Hunan Province (Grant Number 2017JJ1001), and the Advanced Research Program (No. 41412050202).

\section{Reference}
\label{sec:reference}


\begin{thebibliography}{99}
	
	\bibitem{geng2018learning}Geng M, Zhou X, Ding B, et al. Learning to cooperate in decentralized multi-robot exploration of dynamic environments. International Conference on Neural Information Processing. Springer, Cham, 2018: 40-51.
	
	
	\bibitem{Higgins2009survey}Higgins F, Tomlinson A, Martin K M. Survey on security challenges for swarm robotics. 2009 Fifth International Conference on Autonomic and Autonomous Systems. IEEE, 2009: 307-312.
	
	\bibitem{Dresner2008mutliagent}Dresner K, Stone P. A multiagent approach to autonomous intersection management. Journal of artificial intelligence research, 2008, 31: 591-656.
	
	\bibitem{pipa2009multi}Pipattanasomporn, M., Feroze, H., Rahman, S. (2009, March). Multi-agent systems in a distributed smart grid: Design and implementation. In 2009 IEEE/PES Power Systems Conference and Exposition (pp. 1-8). IEEE.
	
	\bibitem{geng2019learning}Geng M, Xu K, Zhou X, et al. Learning to cooperate via an attention-based communication neural network in decentralized multi-robot exploration. Entropy, 2019, 21(3): 294.
	
	
	\bibitem{Millar2013towards}Millard A G, Timmis J, Winfield A F T. Towards exogenous fault detection in swarm robotic systems. Conference towards Autonomous Robotic Systems. Springer, Berlin, Heidelberg, 2013: 429-430.
	
	
	\bibitem{Kilinc18multi} Kilinc, Ozsel, and Giovanni Montana. "Multi-agent deep reinforcement learning with extremely noisy observations." arXiv preprint arXiv:1812.00922 (2018).
	
	\bibitem{li18attention}Iqbal S, Sha F. Actor-attention-critic for multi-agent reinforcement learning. arXiv preprint arXiv:1810.02912, 2018.
	
	\bibitem{Foerster2018counter} Foerster J N, Farquhar G, Afouras T, et al. Counterfactual multi-agent policy gradients. Thirty-Second AAAI Conference on Artificial Intelligence. 2018.
	
	\bibitem{Haarnoja2018Soft} Haarnoja T, Zhou A, Abbeel P, et al. Soft actor-critic: Off-policy maximum entropy deep reinforcement learning with a stochastic actor. arXiv preprint arXiv:1801.01290, 2018.
	
	\bibitem{Kingma2014adam}Kingma D P, Ba J. Adam: A method for stochastic optimization. arXiv preprint arXiv:1412.6980, 2014.
	
	\bibitem{Li15conti}Lillicrap T P, Hunt J J, Pritzel A, et al. Continuous control with deep reinforcement learning. arXiv preprint arXiv:1509.02971, 2015.
	
	\bibitem{Uhlenbeck1930on}Uhlenbeck G E, Ornstein L S. On the theory of the Brownian motion. Physical review, 1930, 36(5): 823.
	
	
	\bibitem{Lowe17multi}Lowe R, Wu Y, Tamar A, et al. Multi-agent actor-critic for mixed cooperative-competitive environments. Advances in Neural Information Processing Systems. 2017: 6379-6390.
	
	
\end{thebibliography}
\end{document}